\pdfoutput=1

\documentclass[11pt]{article}

\usepackage[final]{EMNLP2023}

\usepackage{times}
\usepackage{latexsym}

\usepackage[T1]{fontenc}

\usepackage[utf8]{inputenc}

\usepackage{microtype}

\usepackage{inconsolata}
\usepackage{graphicx}

\title{Context-aware and Style-related Incremental Decoding framework for Discourse-Level Literary Translation}

\author{
  Yuanchang Luo, Jiaxin Guo, Daimeng Wei, Hengchao Shang, Zongyao Li,    \\
  \textbf{
   Zhanglin Wu , Zhiqiang Rao, Shaojun Li, Jinlong Yang, Hao Yang
  }\\
  Huawei Translation Service Center, Beijing, China\\
  \{luoyuanchang1,guojiaxin1,weidaimeng,shanghengchao,lizongyao,\\
 wuzhanglin2,raozhiqiang,lishaojun18,yangjinlong7,yanghao30\}@huawei.com \\
  }

\begin{document}
\maketitle
\begin{abstract}

This report outlines our approach for the WMT24 Discourse-Level Literary Translation Task, focusing on the Chinese-English language pair in the Constrained Track. Translating literary texts poses significant challenges due to the nuanced meanings, idiomatic expressions, and intricate narrative structures inherent in such works. To address these challenges, we leveraged the Chinese-Llama2 model, specifically enhanced for this task through a combination of Continual Pre-training (CPT) and Supervised Fine-Tuning (SFT). Our methodology includes a novel Incremental Decoding framework, which ensures that each sentence is translated with consideration of its broader context, maintaining coherence and consistency throughout the text. This approach allows the model to capture long-range dependencies and stylistic elements, producing translations that faithfully preserve the original literary quality. Our experiments demonstrate significant improvements in both sentence-level and document-level BLEU scores, underscoring the effectiveness of our proposed framework in addressing the complexities of document-level literary translation.
\end{abstract}

\section{Introduction}

Machine Translation (MT) \cite{DBLP:journals/corr/abs-2005-14165,DBLP:journals/jmlr/ChowdheryNDBMRBCSGSSTMRBTSPRDHPBAI23,touvron2023llama} has become an essential tool in breaking language barriers, enabling the automatic translation of text from one language to another. While significant advancements \cite{DBLP:conf/nips/VaswaniSPUJGKP17,sennrich2016improving,wei2023text,DBLP:conf/iclr/Gu0XLS18,DBLP:conf/emnlp/GhazvininejadLL19,DBLP:conf/inlg/WangGWCSWZTY21,DBLP:journals/corr/abs-2112-11640,DBLP:journals/corr/abs-2112-11642} have been made in MT for various text genres, translating literary texts remains a formidable challenge. Literary texts are rich in complex linguistic phenomena, such as nuanced meanings, idiomatic expressions, and intricate narrative structures. Unlike technical or news-related texts, literary works demand a deeper understanding of context, tone, and style, making them particularly challenging for MT systems. This difficulty is compounded by the scarcity of high-quality parallel datasets in the literary domain, limiting the ability of MT models to learn from extensive, diverse examples.

\begin{figure*}
\centering
\includegraphics[width=15cm]{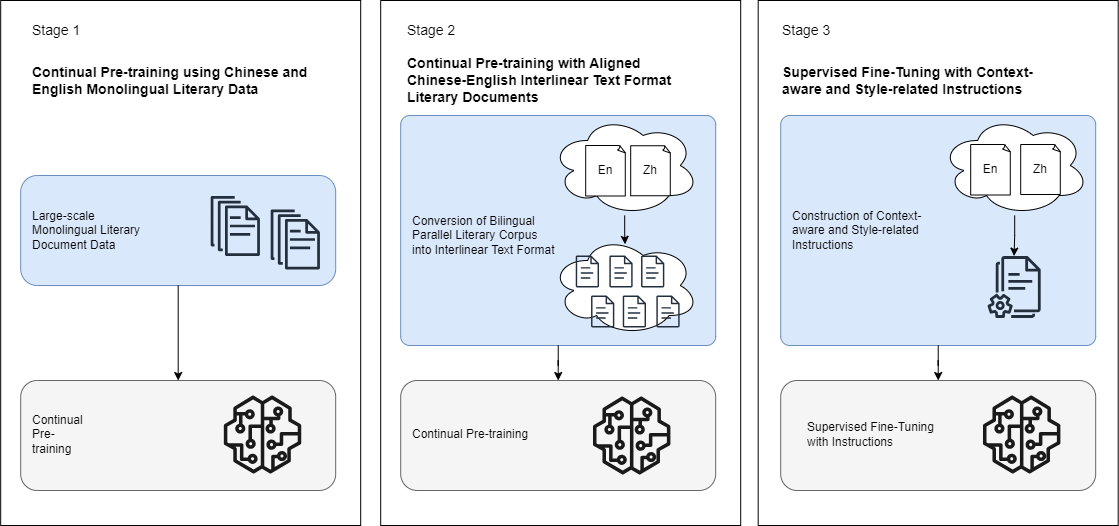}
\caption{The overall of our approach.}
\label{figure:model}
\end{figure*}

Document-level translation \cite{Sun_Wang_Zhou_Zhao_Huang_Chen_Li_2020,duextrapolation,wu2024perhaps} introduces another layer of complexity to MT, especially when dealing with longer texts such as novels. Unlike sentence-level translation, where context is limited to a single sentence, document-level translation requires the model to consider the broader discourse context to maintain coherence and consistency throughout the entire text. This is particularly crucial in literary translation, where the narrative thread, character development, and thematic elements must be preserved across sentences and paragraphs. Long-range dependencies, where information introduced early in a text influences later parts, pose a significant challenge for MT systems, which often struggle to retain and apply such context effectively over extended texts.

In this system report, we describe our participation in the WMT24 Discourse-Level Literary Translation Task, focusing on the Chinese-English language pair under the Constrained Track. Our approach leverages the Chinese-Llama2 model, specifically designed for this task, through a combination of Continual Pre-training (CPT) and Supervised Fine-Tuning (SFT). This methodology allows us to refine the model's understanding of literary texts while adapting it to the specific nuances of Chinese-English translation. Additionally, we employ an Incremental Decoding framework, which enables the model to translate documents sentence by sentence, ensuring that each translation is informed by the broader context. This approach is designed to tackle the challenges of document-level literary translation, aiming to produce translations that are not only accurate but also faithful to the original text's literary quality.

\section{Background: TP3}

Machine Translation (MT) is the automated process of converting text from one language to another using computational methods. Traditionally, MT relies on encoder-decoder models, where the encoder processes the source language and the decoder generates the translation, often requiring large bilingual datasets and data augmentation to improve performance. Recently, Large Language Models (LLMs) like GPT have become prominent in MT, enabling translation through zero-shot or few-shot learning by conditioning on a source sentence \cite{jiao2023parrot,zeng2023tim,chen2023improving,xu2023paradigm,yang2023bigtranslate,zhang2023bayling}. These models can also be fine-tuned with high-quality bilingual data and tailored instructions to enhance translation accuracy and robustness, offering new possibilities for MT with limited resources.

\paragraph{TP3}

\citet{DBLP:conf/naacl/GuoYLWSC24} propose a novel training paradigm, consisting of Three-Stages Translation Pipeline (TP3), to boost the translation capabilities of LLMs. The training paradigm includes:

Stage 1: Continual Pre-training using Extensive Monolingual Data.
This stage aims to expand the multilingual generation capabilities of LLMs. While it is inherently related to machine translation tasks, it is not essential.

Stage 2: Continual Pre-training with Interlinear Text Format Documents.
They construct interlinear text format from sentence-aligned bilingual parallel data and utilize them for continual pre-training of LLMs. Experimental results demonstrate the critical importance of this stage, resulting in a significant improvement in translation quality, particularly for English-Other translations. 

Stage 3: Leveraging Source-Language Consistent Instruction for Supervised Fine-Tuning. 
In this stage, they discover that setting instructions consistent with the source language benefits the supervised fine-tuning process.

\section{Methods}

\subsection{TP3 for Discourse-Level Literary}

We introduce the TP3 training paradigm into the literary translation task, with the entire training process illustrated in Figure \ref{figure:model}.

\paragraph{Stage 1: Continual Pre-training using Chinese and English Monolingual Literary Data}



\begin{figure*}
\centering
\includegraphics[width=15cm]{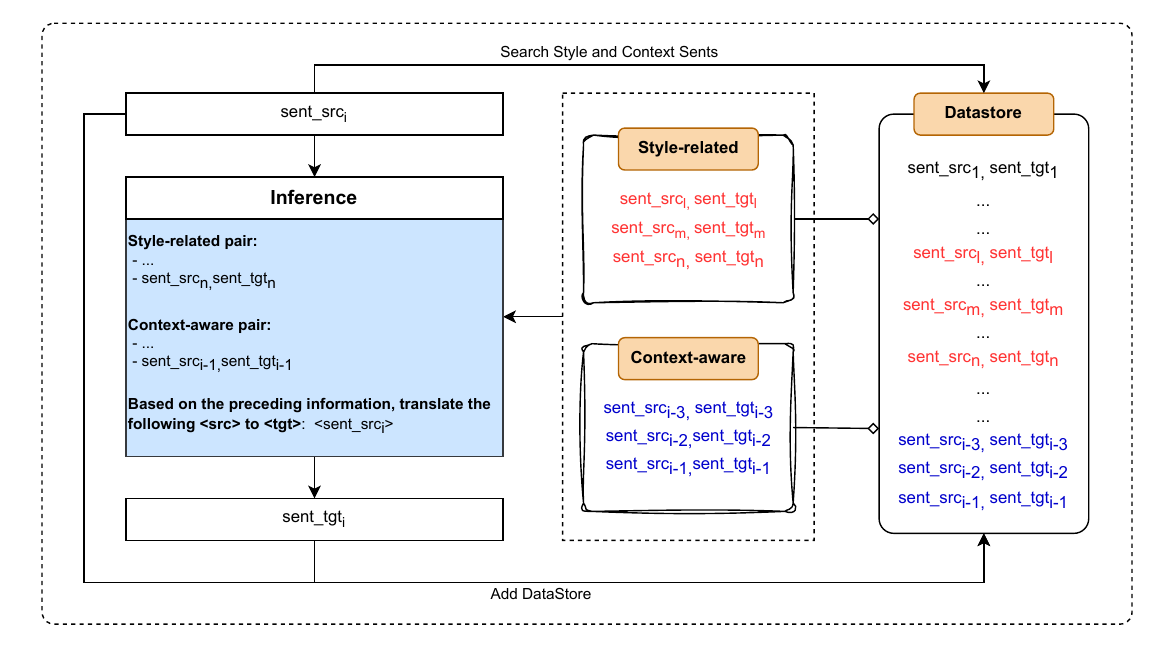}
\caption{The overall of our incremental decoding framework.}
\label{figure:decoding}
\end{figure*}

In this stage, we adapt a general-purpose large language model (LLM) into a specialized Literary LLM by using monolingual literary data in both Chinese and English. While existing LLMs like Llama perform well in English-centric tasks, their capabilities in other languages, especially in literary contexts, are often limited. To improve this, we employ continual pre-training with extensive monolingual literary texts, enhancing the model's understanding of nuanced language, stylistic elements, and narrative structures. This step is critical for enabling the model to generate more coherent and contextually appropriate translations.

For this task, continual pre-training is essential, transforming a general LLM into one tailored for literary translation. We treat each novel as a distinct training unit, combining sentences within each chapter into paragraphs to capture long-range dependencies and context. This approach is vital for maintaining consistency and preserving the literary quality of translations. By focusing on both Chinese and English literary data, the model gains a balanced understanding of the stylistic and structural intricacies in both languages.

\paragraph{Stage 2: Continual Pre-training with Aligned Chinese-English Interlinear Text Format Literary Documents}

In Stage 2, we enhance the model's cross-lingual translation capabilities by using aligned Chinese-English interlinear text format literary documents, building on the foundation established in Stage 1. The interlinear text format, where each source sentence is directly aligned with its translation at the word or phrase level, is essential for enabling the model to understand and map the syntactic and semantic structures between Chinese and English, which is crucial for producing high-quality translations. We implement a continual pre-training approach using LoRA (Low-Rank Adaptation of Large Language Models) \cite{hu2021lora} to efficiently adapt the model with these interlinear text documents.

Initially, the model was trained on general sentence-aligned parallel data to establish a strong cross-lingual alignment foundation. Subsequently, we performed incremental pre-training with literary-specific interlinear data. By focusing on literary documents, we ensure the model becomes finely attuned not only to general cross-lingual translation but also to the unique stylistic and structural nuances of literary texts. This approach enables the model to capture the intricate relationships between Chinese and English in a literary context, significantly improving translation quality and fidelity.

\paragraph{Stage 3: Supervised Fine-Tuning with Context-aware and Style-related Instructions}

In the final stage of our approach, we conduct supervised fine-tuning using context-aware and style-related instructions, specifically tailored to address the challenges of semantic coherence and stylistic consistency in literary translation. Unlike the traditional approach of using Source-Language Consistent Instruction, which emphasizes alignment with the source language, our method focuses on ensuring that the translated output maintains a consistent narrative flow and adheres to the stylistic nuances of the original text. This adjustment is crucial for literary translation, where preserving the author's voice and the overall tone of the work is just as important as achieving accurate translation.

The fine-tuning process leverages the LoRA to refine specific parameters of the model efficiently. By applying LoRA, we can update the model with low-rank adaptations, which helps in preventing overfitting while ensuring that the model adapts effectively to the task-specific requirements. This targeted fine-tuning allows the model to better capture the long-range dependencies and stylistic elements that are essential for producing translations that are not only accurate but also faithful to the literary qualities of the source text.

\subsection{Incremental Decoding framework}

\begin{table*}[th]
\centering
\resizebox{0.98\linewidth}{!}{
\begin{tabular}{lcccccccc}
\hline
 & \multicolumn{2}{c}{\textbf{Valid 1}} & \multicolumn{2}{c}{\textbf{Valid 2}} & \multicolumn{2}{c}{\textbf{Test 1}} &  \multicolumn{2}{c}{\textbf{Test 2}}\\
\cline{2-9}
 & s-BLEU & d-BLEU & s-BLEU & d-BLEU & s-BLEU & d-BLEU & s-BLEU & d-BLEU \\
\hline
General Sent-Trans & 16.81 & 24.1 & 10.74 & 17.39 & 17.97 & 25.87 & 13.32 & 20.37 \\
Literary Sent-Trans & 23.35 & 30.51 & 14.64 & 21.81 & 20.91 & 28.51 & 18.02 & 25.38 \\
Literary Doc-Trans & 23.78 & 31.85 & 14.94 & 22.12 & 20.97 & 29.43 & 18.28 & 25.62 \\
\hline
\end{tabular}
}
\caption{\textbf{The overall results.}}
\label{table:main-results}
\end{table*}

In traditional machine translation, sentences are often translated independently of one another, leading to issues with semantic coherence and stylistic consistency when viewed from a broader, document-level perspective. To address these challenges, we propose an Incremental Decoding framework that considers the translation of each sentence as part of a continuous process, taking into account the translations of previous sentences. This method ensures that the translated text maintains a cohesive flow and consistent style throughout the entire document.

The Incremental Decoding framework incorporates two key components: Context-aware information and Style-related information. Context-aware information involves using the translations of the previous n sentences as historical context when translating the current sentence. This helps maintain continuity in the narrative and ensures that the translation aligns with the broader context established in earlier sentences.

Style-related information further refines this process by incorporating translations of sentences that are similar to the current sentence in terms of content and style. These sentences are selected based on sentence and keyword similarity, ensuring that the translation reflects the stylistic nuances present in the original text. By integrating both context-aware and style-related information, the Incremental Decoding framework produces translations that are not only accurate but also harmonious in tone and structure, closely mirroring the original literary work.

\section{Experiments}

\subsection{Datasets and Evaluation Metrics}

We utilized data from the general MT shared task and the GuoFeng Webnovel Corpus. The GuoFeng Webnovel Corpus was employed in Stages 1, 2, and 3, while the general MT data was used exclusively in Stage 2. Detailed statistics of the data are presented in Table \ref{tb:data}.

\begin{table}[ht]
\begin{center}
\begin{tabular}{@{}cc@{}}
\hline
Data Source &  Data Size \\
\hline
General MT & 25M \\
GuoFeng Webnovel Corpus & 1.9M \\
\hline
\end{tabular}
\caption{Data Statistics.}
\label{tb:data}
\end{center}
\end{table}

For the evaluation metrics, we utilized SacreBLEU \cite{DBLP:conf/acl/PapineniRWZ02} to assess system performance. Given that the test set was segmented into sentence-level units, we conducted evaluations using both s-BLEU (sentence-level BLEU) and d-BLEU (document-level BLEU) scores to provide a comprehensive analysis of the translation quality.

\subsection{Experiment Settings}

In our experiments, we used Chinese-LLaMA2 \cite{Chinese-LLaMA-Alpaca} as the foundation model. Chinese-LLaMA2 is an enhanced and optimized version of Llama-2, specifically designed for Chinese language understanding and instruction comprehension. This model includes a larger Chinese vocabulary and benefits from incremental pretraining on a large-scale Chinese dataset, which significantly improves its semantic understanding capabilities.

For both the Continual Pre-training and Supervised Fine-Tuning stages, we adhered to the hyperparameters utilized in the Chinese-LLaMA2 project. During Stage 2, the model was trained for 1 epoch, while in Stage 3, the training was extended to 3 epochs to ensure more refined adjustments.

All experiments were conducted using 8 Nvidia GPUs, each with 64GB of memory, and employed DeepSpeed \cite{DBLP:conf/kdd/RasleyRRH20} ZeRO 2 for model parallelization, which allowed for efficient handling of the large-scale model and dataset.

\subsection{Compared Baselines}

\begin{itemize}
    \item \textbf{General Sent-Trans}: In this baseline, we directly create sentence-level translation instruction data and use it to perform Supervised Fine-Tuning on the Chinese-LLaMA2 model. This approach focuses on training the model with general sentence-level translation tasks without any specialized pre-training.
    \item \textbf{Literary Sent-Trans}: This baseline builds on the previous stages, as outlined in Stage 1 and Stage 2. We first subject the Chinese-LLaMA2 model to Continual Pre-training using monolingual and bilingual literary data. Following this pre-training, the model undergoes Supervised Fine-Tuning using the same sentence-level translation instruction data as in the General Sent-Trans baseline. This approach is designed to adapt the model to the literary domain before fine-tuning it with general sentence-level instructions.
    \item \textbf{Literary Sent-Trans}: \textbf{This represents our final proposed approach}. After the Continual Pre-training conducted in Stage 1 and Stage 2, we further train the model using the Supervised Fine-Tuning method from Stage 3, which incorporates Context-aware and Style-related Instructions. This method aims to enhance the model's ability to maintain semantic coherence and stylistic consistency across sentences in literary document translation.
\end{itemize}

\subsection{Results}

The comparison between Literary Sent-Trans and General Sent-Trans reveals significant improvements in both s-BLEU and d-BLEU scores across various test sets, indicating that Stage 1 and Stage 2 effectively incorporated literary knowledge into the model. Furthermore, when comparing Literary Doc-Trans with Literary Sent-Trans, we observe additional gains in both s-BLEU and d-BLEU metrics, demonstrating the effectiveness of Stage 3's Context-aware and Style-related Instructions. These results collectively highlight the incremental benefits of each stage in enhancing the model's performance in literary translation. The detailed results are presented in Table \ref{table:main-results}.

\section{Conclusion}

In this work, we addressed the complex task of literary translation within the WMT24 Discourse-Level Literary Translation Task, focusing on the Chinese-English language pair. By leveraging the Chinese-Llama2 model, enhanced through Continual Pre-training and Supervised Fine-Tuning, we successfully adapted the model to capture the unique nuances of literary texts. Our Incremental Decoding framework further ensured that each sentence was translated with awareness of its broader context, resulting in more coherent and stylistically consistent translations. The improvements observed in both sentence-level and document-level BLEU scores validate the effectiveness of our approach. These results highlight the potential of combining advanced language models with specialized training strategies to tackle the intricacies of literary translation, paving the way for further research in this challenging domain.

\bibliography{emnlp2023}
\bibliographystyle{acl_natbib}

\end{document}